\pdfoutput=1
\documentclass[11pt]{article}

\usepackage{EMNLP2022}

\usepackage{times}
\usepackage{latexsym}

\usepackage[T1]{fontenc}
\usepackage[utf8]{inputenc}

\usepackage{microtype}

\usepackage[utf8]{inputenc} 
\usepackage[T1]{fontenc}    
\usepackage{hyperref}       
\usepackage{url}            
\usepackage{booktabs}       
\usepackage{amsfonts}       
\usepackage{nicefrac}       
\usepackage{xcolor}
\usepackage{epsfig}
\usepackage{algorithm}
\usepackage{algorithmic}
\usepackage{amsthm}
\usepackage{amsmath}
\usepackage{amssymb}
\usepackage{microtype}
\usepackage{mathtools}
\usepackage{float}
\usepackage{graphicx}
\usepackage{wrapfig}
\usepackage{makecell}
\usepackage{caption}
\usepackage{subcaption}
\usepackage{bm}
\usepackage{bbm}
\usepackage{multirow}
\usepackage{microtype}
\usepackage{mathtools}
\usepackage{enumitem}
\usepackage{cancel}
\usepackage{chngcntr}
\usepackage{wrapfig}
\usepackage{cutwin}
\usepackage{lipsum}

\newcommand{\mask}{\ttt{[MASK]}}
\newcommand{\cls}{\ttt{[CLS]}}

\newcommand{\ba}{$_\texttt{base}$}

\DeclareMathOperator*{\softmax}{softmax}
\DeclareMathOperator*{\attention}{Attention}
\DeclareMathOperator{\var}{Var}

\newcommand\tf[1]{\textbf{#1}}
\newcommand\ttt[1]{\texttt{#1}}

\title{Differentiable Data Augmentation \\for Contrastive Sentence Representation Learning}

\author{Tianduo Wang \and Wei Lu\\
  StatNLP Research Group\\
  Singapore University of Technology and Design \\
  \texttt{\{tianduo\_wang,luwei\}@sutd.edu.sg} \\}

\begin{document}
\maketitle

\begin{abstract}
Fine-tuning a pre-trained language model via the contrastive learning framework with a large amount of unlabeled sentences or labeled sentence pairs is a common way to obtain high-quality 
sentence representations.
% Ok
Although the contrastive learning framework has shown its superiority on sentence representation learning over previous methods, the potential of such a framework is under-explored so far due to the simple method it used to construct positive pairs.
% Ok
Motivated by this, we propose a method that makes {\em hard positives} from the original training examples.
% Ok
A pivotal ingredient of our approach is the use of prefix that is attached to a pre-trained language model, which allows for {\em differentiable} data augmentation during contrastive learning.
% Ok
%
Our method can be summarized in two steps: {\em supervised prefix-tuning} followed by {\em joint contrastive fine-tuning} with unlabeled or labeled examples.  
% Ok
Our experiments confirm the effectiveness of our data augmentation approach.
% Ok
The proposed method yields significant improvements over existing methods under both semi-supervised and supervised settings. 
Our experiments under a low labeled data setting also show that our method is more label-efficient than the state-of-the-art contrastive learning methods.
\footnote{Our code and model checkpoints are available at \url{https://github.com/TianduoWang/DiffAug}.}
% Not good

\end{abstract}

\section{Introduction}

Learning universal and effective sentence representations is an enduring problem in natural language processing (NLP). 
The objective of learning sentence representations is similar to that of learning word embeddings~\cite{mikolov2013skipgram,pennington2014glove}, i.e., we expect the embeddings of sentences with similar semantics to be close to each other, while the embeddings of sentences with different meanings to be sufficiently far away from each other.
Recently, we have witnessed the success of pre-trained language models in various NLP tasks.
However, the quality of the sentence representations directly obtained from non fine-tuned language models remains \textcolor{black}{unsatisfactory}~\cite{reimers2019sbert,li2020bertflow}.

One approach to addressing this problem is to fine-tune a pre-trained language model via the contrastive learning objective on labeled or unlabeled sentences.
Several recent research efforts~\cite{yan2021consert,gao2021simcse,jiang2022promptbert} have shown that a contrastive learning based fine-tuning stage can produce state-of-the-art results on the sentence embedding learning task.

Some previous contrastive learning works on other modalities, e.g., image~\cite{chen2020simclrv1,chen2020simclrv2},
have shown that an effective data augmentation (DA) method is crucial for the success of contrastive learning. 
However, due to the discrete nature of language, 
\textcolor{black}{applying commonly-used sentence augmentation strategies,}
e.g., word deletion and replacement, over the input sentences for contrastive learning leads to suboptimal results~\cite{gao2021simcse}. 
Instead, \citet{gao2021simcse} propose to perform data augmentation via dropout, showing that this extremely simple approach can yield better sentence representations than previous DA methods that are based on discrete transformations.

\begin{figure}[t]
    \centering
    \includegraphics[width=0.9\linewidth]{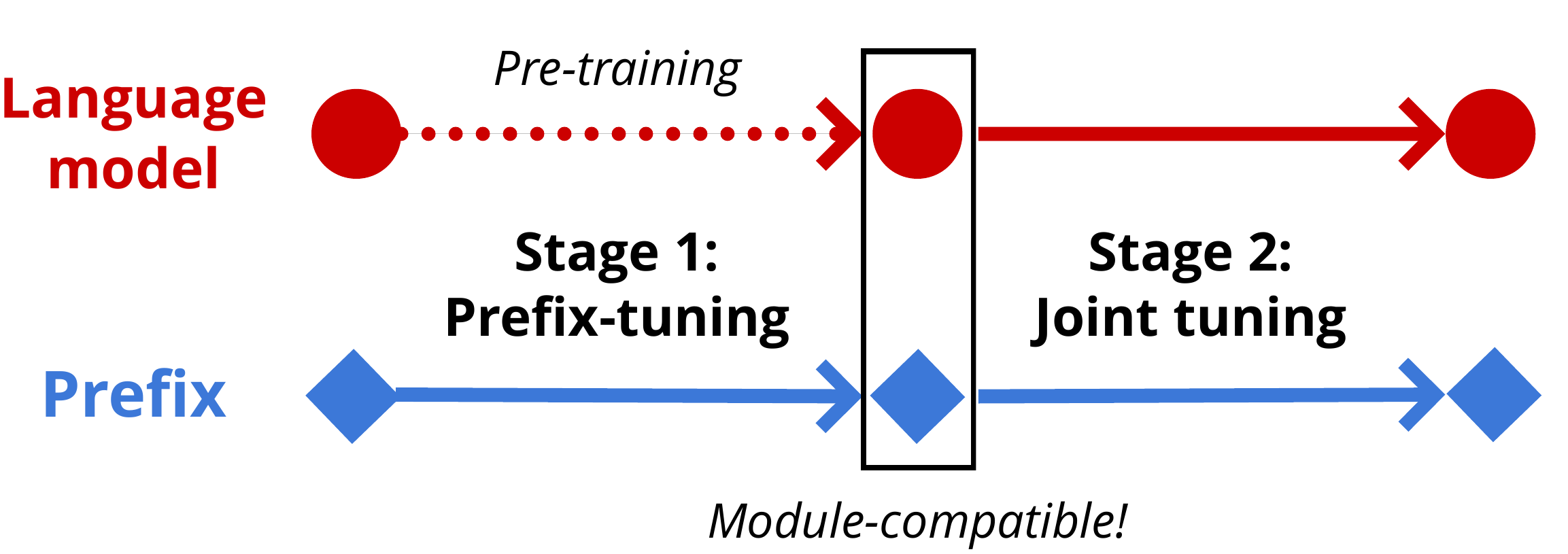}
    \vspace{-5pt}
    \caption{\label{fig: tt2} Our two-stage tuning strategy.}
\end{figure}

Although this \textcolor{black}{dropout-based} DA method outperforms its discrete counterparts, the potential of the contrastive learning framework is under-explored due to its simple treatment of the positive pair construction process.
Previous works have shown that contrastive learning benefits from strong data augmentations that can produce hard positives with meaningful differences.
Specifically, \citet{chen2020simclrv1} demonstrate that contrastive learning requires stronger data augmentation than traditional supervised learning.
\citet{tian2020infomin} further show that, for contrastive learning, a good pair of positive instances should only share task-relevant information while discarding irrelevant information as much as possible.

Motivated by this idea, we propose \tf{DiffAug}, a \tf{\underline{diff}}erentiable data \tf{\underline{aug}}mentation method for contrastive sentence representation learning. 
This method prepends two separate prefix modules~\cite{li2021prefix} to a common pre-trained language model. 
The goal is to obtain hard positive pairs with the help of these two prefix modules for contrastive learning, where supervised signals can be used to guide the tuning process of such prefix modules. 
Such a design essentially requires us to fine-tune both language model and newly-added prefix.
However, how to effectively train them jointly to best serve our sentence representation learning purpose is a non-trivial research question.

Our observations show that, though tempting, it can be undesirable to fine-tune language model and prefix jointly from the beginning of the training process.
Rather, we propose an effective two-stage tuning strategy, 
{\em prefix-tuning} followed by {\em joint tuning}, 
as illustrated in Figure~\ref{fig: tt2}. 
Specifically, we argue that it is crucial to perform prefix-tuning first, until a {\em module-compatible} state is achieved, before we move to the second stage to perform joint tuning for both modules. 
Our main contributions can be summarized as follows:
\begin{itemize}[topsep=0pt, partopsep=0pt, leftmargin=15pt, parsep=0pt, itemsep=5pt]
    
    \item We propose a novel and effective data augmentation method for contrastive sentence representation learning 
    with the help of prefix modules.
    We further design a mechanism that allows this module to be carefully optimized with labeled data first, allowing hard positives with meaningful differences to be constructed, which can benefit contrastive learning.

    \item Our experiments show that the proposed method achieves the new state-of-the-art results on both semi-supervised and supervised settings. We also investigate the situation when labeled data is scarce. 
    The results demonstrate that our method is more label-efficient than previous contrastive learning methods, showing the robustness of our proposed approach.

    \item 
    \textcolor{black}{Our work successfully combines the traditional fine-tuning and prefix-tuning paradigms.}
    Through extensive analysis we identify the crucial elements required to ensure the success of our proposed approach.

\end{itemize}

\section{Background}

Our method is based on the state-of-the-art contrastive learning framework proposed by~\citet{gao2021simcse}.
We first introduce this framework.
Next, we discuss the mechanism of prefix-tuning~\cite{li2021prefix}.

\subsection{The contrastive learning framework}

Contrastive learning aims to learn meaningful representations from data by pulling together instances with similar semantic meanings and pushing apart dissimilar ones~\cite{hadsell2006dimensionality}. 
The contrastive learning framework that we apply in this work is proposed by~\citet{gao2021simcse}.
There are three main components in this framework:

\begin{itemize}[topsep=0pt, partopsep=0pt, leftmargin=15pt, parsep=0pt, itemsep=5pt]

    \item A data augmentation module $g(\cdot\ ;\theta)$ that generates positive pairs from a batch of training data with parameters denoted as $\theta$. 
    Previous contrastive learning works on other modalities~\cite{chen2020simclrv1,tian2020infomin} have shown that the meaningful differences between positive pairs will allow the potential of contrastive learning to be better explored. Our work mainly improves this module.

    \item A neural network based encoder $f(\cdot\ ; \phi)$ that maps input data to a representation space with parameters denoted as $\phi$.
    For natural language, $f(\cdot\ ; \phi)$ could be any neural network architecture that is suitable for encoding sentences.
    For our approach, we follow previous works~\cite{gao2021simcse,jiang2022promptbert}, and employ pre-trained language models as $f(\cdot\ ; \phi)$, e.g. BERT~\cite{devlin2019bert} and RoBERTa~\cite{liu2019roberta}.\footnote{%
    One common method to obtain sentence embedding is to use the \cls \ representation of the pre-trained language model.
    In this work, we add a hard prompt with a mask token \mask \ to each input sentence and obtain sentence embeddings from the \mask \ representations, following PromptBERT~\cite{jiang2022promptbert}.}
    
    \item A contrastive loss function can be defined over a batch of data with size $N$ as follows:
    \begin{equation}
        \mathcal{L}_{\text{cl}} = - \frac{1}{N} \sum_{i=1}^{N} \log\frac{\exp(\frac{\cos(h_{i,1},\ h_{i,2})}{\tau})}{\sum_{j=1}^N \exp(\frac{\cos(h_{i,1},\ h_{j,2})}{\tau})}
    \end{equation}

    where $h_{i,1} = f(g(x_i;\theta_1); \phi)$ and $h_{i,2} = f(g(x_i;\theta_2); \phi)$ are the representations of the augmented instances from the same data $x_i$, 
    $\tau$ is the temperature hyperparameter, 
    and $\cos(\cdot, \cdot)$ is the cosine similarity function. 
    Note that we have two parameters $\theta_1$ and $\theta_2$, as we involve two separate data augmentation operations here.
\end{itemize}

\subsection{The mechanism of prefix-tuning}

Transformer~\cite{vaswani2017attention} has become the most crucial component for many pre-trained language models. In this section, we discuss the self-attention module in Transformer and illustrate how prefix-tuning~\cite{li2021prefix} works. 
The original self-attention layer first maps the input $X \in \mathbb{R}^{L \times d}$ into three matrices, i.e., query $Q\in \mathbb{R}^{L \times d}$, key $K\in \mathbb{R}^{L \times d}$, and value $V\in \mathbb{R}^{L \times d}$, where $L$ and $d$ are the input length and hidden dimension respectively.
The self-attention function is defined as:
\begin{equation}
H = \attention(Q, K, V) = \softmax(\frac{QK^T}{\sqrt{d}})V
\end{equation}
where $H \in \mathbb{R}^{L \times d}$ is the output of the self-attention layer. Prefix-tuning~\cite{li2021prefix} affects the output of the original language model by prepending {\em tunable} matrices as key-value pairs. Specifically, in each self-attention layer, two matrices $P_k,\ P_v \in \mathbb{R}^{l \times d}$ will be concatenated to the original $K$ and $V$ respectively, where $l$ is the prefix length. Therefore, in prefix-tuning, the self-attention function will become the following:
\begin{equation}
\begin{aligned}
H_p &= \attention(Q, [P_k;K], [P_v;V]) \\&= \softmax(\frac{Q[{P}_k;K]^T}{\sqrt{d}})[P_v;V]
\end{aligned}
\end{equation}

\noindent
where $[\cdot ; \cdot]$ represents the concatenation on the first dimension.\footnote{We have $[{P}_k;{K}] , \ [{P}_v;{V}] \in \mathbb{R}^{(l+L) \times d}$, and ${H}_p \in \mathbb{R}^{L \times d}$ is the output of self-attetion layer in prefix-tuning.}
Since our method applies prefix for data augmentation, we use $\theta$ to denote the set of {\em tunable} matrices $\{(P_k^i, P_v^i)|i=1,...,M\}$, where $M$ is the number of self-attention layers in the pre-trained language model.

\section{Method}

In this section, we describe the proposed data augmentation method, and our two-stage tuning strategy.
We then discuss reasons why our proposed approach works.

\subsection{Differentiable data augmentation}
\label{sec: diffaug}

The InfoMin principle~\cite{tian2020infomin} states that a good pair of positive instances for contrastive learning should be as different as possible while retaining enough useful information relevant to the downstream tasks.
Following this principle, we improve the current state-of-the-art contrastive learning framework~\cite{gao2021simcse} via prefix~\cite{li2021prefix}.
We hope the added prefix modules can generate positive pairs with meaningful differences.
To formulate this idea, we replace $h_i = f(g(x_i;\theta);\phi)$ with $h_i = f(x_i; \theta, \phi)$, since the prepended prefix modules can be regarded as a part of the language model.

We initialize the prefix modules randomly following~\citet{li2021prefix}.
This step allows the two initial representations returned from the two networks (with two different prefix modules) to be reasonably different.
The next question is how to inject task-relevant information into the prefix.
Previous works~\cite{conneau2017nlisent,reimers2019sbert} have shown that training encoders with natural language inference (NLI) datasets~\cite{bowman2015snli,williams2018mnli} via cross entropy loss can enhance the quality of generated sentence embeddings.
It motivates us to use this objective to train our prefix modules.
Since $\theta$ is initialized randomly, and $\phi$ is obtained from pre-training, we believe directly tuning $\theta$ and $\phi$ together from the beginning may not lead to the optimal results.
Instead, we propose a two-stage tuning strategy: 
during the first stage, 
we only tune the prefix via cross entropy loss over the NLI dataset with the language model fixed,
while in the second stage, we tune $\theta$ and $\phi$ jointly via the contrastive learning framework.

\paragraph{Stage 1: prefix-tuning.}

\begin{figure}[t]
\centering
\begin{minipage}[c]{1.0\linewidth}
  \centering
  \small
  \begin{subfigure}{.40\textwidth}
    \centering
    \includegraphics[width=\textwidth]{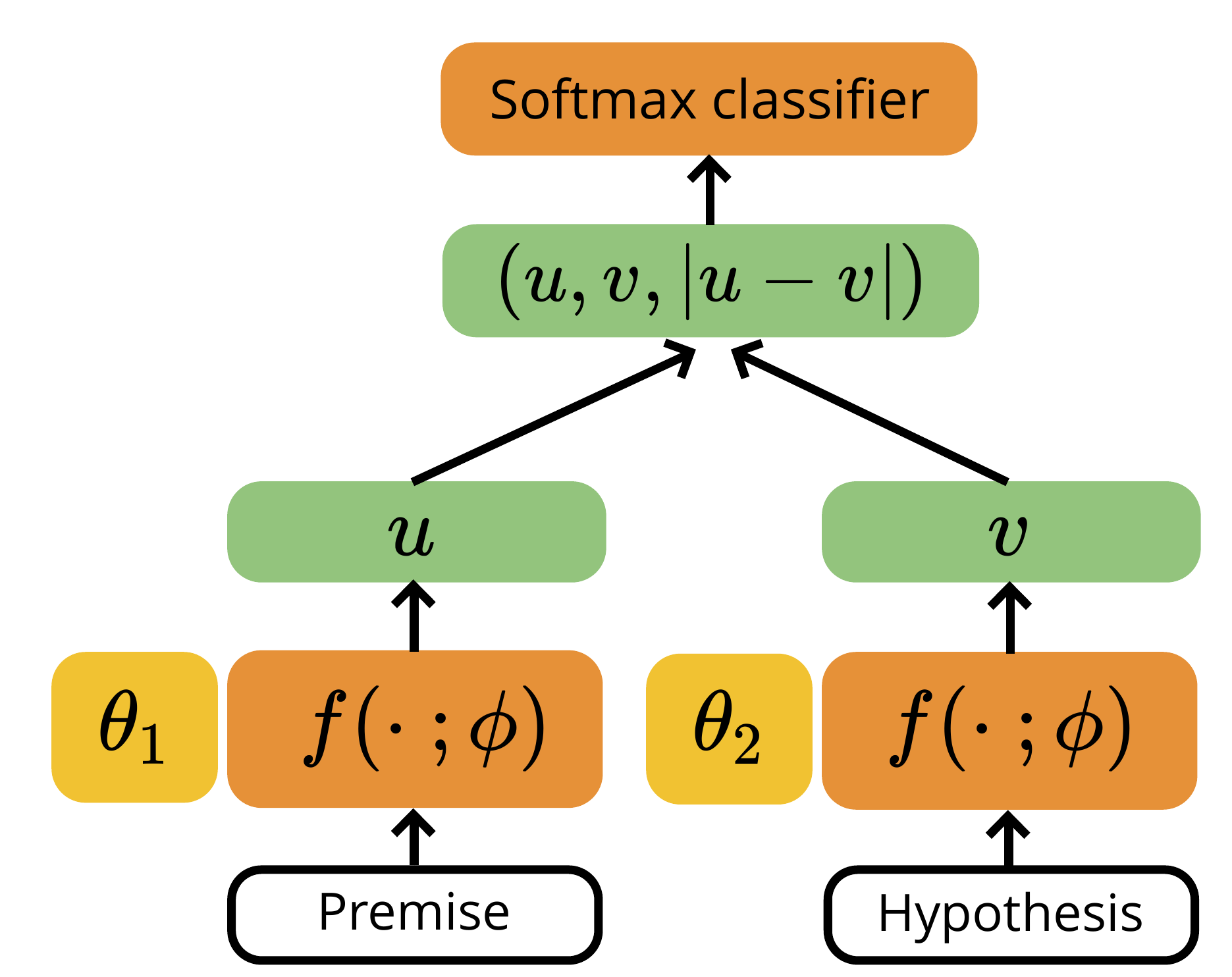}
    \captionsetup{font={scriptsize}}
    \caption{\label{fig:stage1}Stage 1: prefix-tuning}
  \end{subfigure}
  \hspace{+0.05cm}
  \begin{subfigure}{.57\textwidth}
    \centering
    \includegraphics[width=\textwidth]{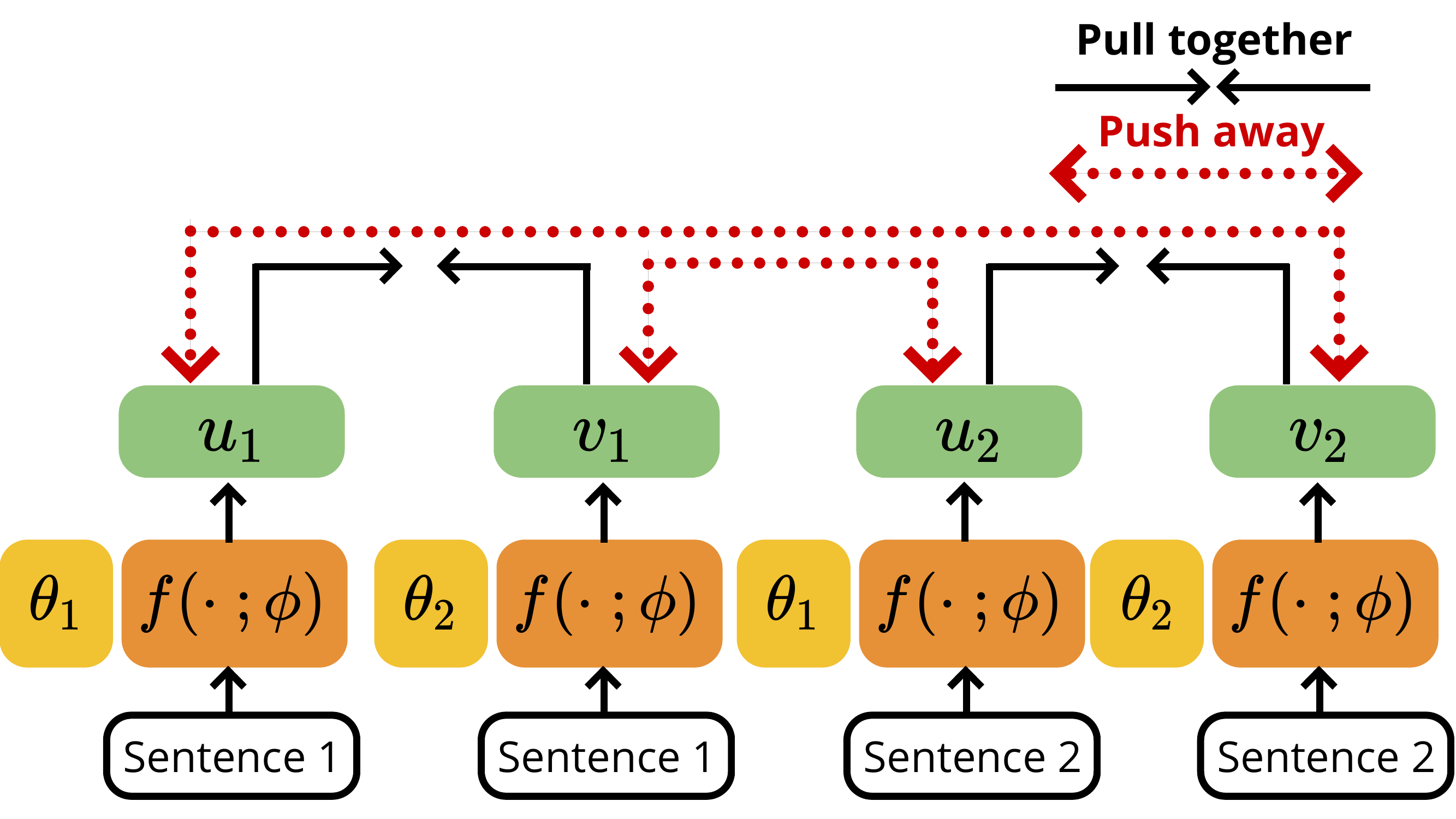}
    \vspace{-11pt}
    \captionsetup{font={scriptsize}}
    \caption{\label{fig:stage2} Stage 2: joint tuning}
  \end{subfigure}
  \caption{\label{fig:framework}The proposed DiffAug method has two training stages: (a) use cross entropy loss to make prefix modules capable of capturing the relationship between sentence pairs with the language model fixed; (b) use the contrastive learning objective to optimize both language model and prefix modules.}
\end{minipage}\hfill
\end{figure}

There are two main objectives for stage-1 tuning.
First, we hope the two prefix modules to have meaningful differences such that they are capable of capturing the relation between NLI sentence pairs.
Second, we expect the added prefix modules to be compatible with the original language model, so that they can be trained jointly in stage 2.
To fulfill these two objectives, we optimize:
\begin{equation}
\min_{\theta_1, \theta_2} \mathcal{L}_{\text{ce}}(f(X_p; \theta_1, \phi), f(X_h; \theta_2, \phi), Y)
\end{equation}
where $X_p$ and $X_h$ represent premise and hypothesis sentences from the NLI dataset respectively, 
and $Y$ consists of binary labels indicating the relation between sentence pairs formed from $X_p$ and $X_h$.
Here $\theta_1$ and $\theta_2$ represent the parameters of two prepended prefix modules respectively.
The feature vector we use for the cross entropy loss is the concatenation of two sentence representations $u$ and $v$, 
as well as their element-wise absolute differences $|u-v|$ following~\cite{reimers2019sbert}. Stage-1 tuning is illustrated in Figure~\ref{fig:stage1}.
\paragraph{Stage 2: joint tuning.} \label{sec:stage2}
After obtaining a pair of well-trained prefix modules from stage 1, we do contrastive learning in stage-2 tuning.
The following objective is optimized:
\begin{equation}
\min_{\theta_1, \theta_2, \phi} \mathcal{L}_{\text{cl}}(f(X; \theta_1, \phi), f(X; \theta_2, \phi))
\end{equation}

Our stage-2 tuning is illustrated in Figure~\ref{fig:stage2}. Previous works~\cite{yan2021consert,li2020bertflow} have shown that adding cross entropy loss on NLI data as an auxiliary loss during unsupervised fine-tuning can enhance the quality of the learned sentence embeddings.
We add this auxiliary loss during stage 2 under our semi-supervised setting, and find it further improves the performance.
More details about the auxiliary loss are in Appendix~\ref{app:aux}.

\subsection{Why two-stage tuning works?} \label{sec:why-work}

    In this section, we explain why the proposed two-stage tuning strategy works.
    From experimental results, we find the performance of contrastive learning is sensitive to the number of training steps in stage 1, and there is an optimal value for this hyperparameter. 
    This phenomenon is consistent for both BERT and RoBERTa.
    To better understand this phenomenon, we introduce a new concept here: {\em module-compatible state}.
    When we say two modules are in the module-compatible state, we mean these two modules can be tuned together while obtaining satisfactory results after training.
    Given a set of training hyperparameters (e.g., learning rate and batch size) and a stage-2 training objective, the extent of module compatibility is closely related to the number of stage-1 training steps.
    We find there are two countering metrics that determine the optimal stage-1 steps: {\em representation divergence} ($\delta$) and {\em weight convergence} ($\kappa$).

\begin{figure}[t]
  \centering
  \begin{minipage}[c]{1.0\linewidth}
    \centering
    \small
    \begin{subfigure}{.48\textwidth}
      \centering
      \includegraphics[width=\textwidth]{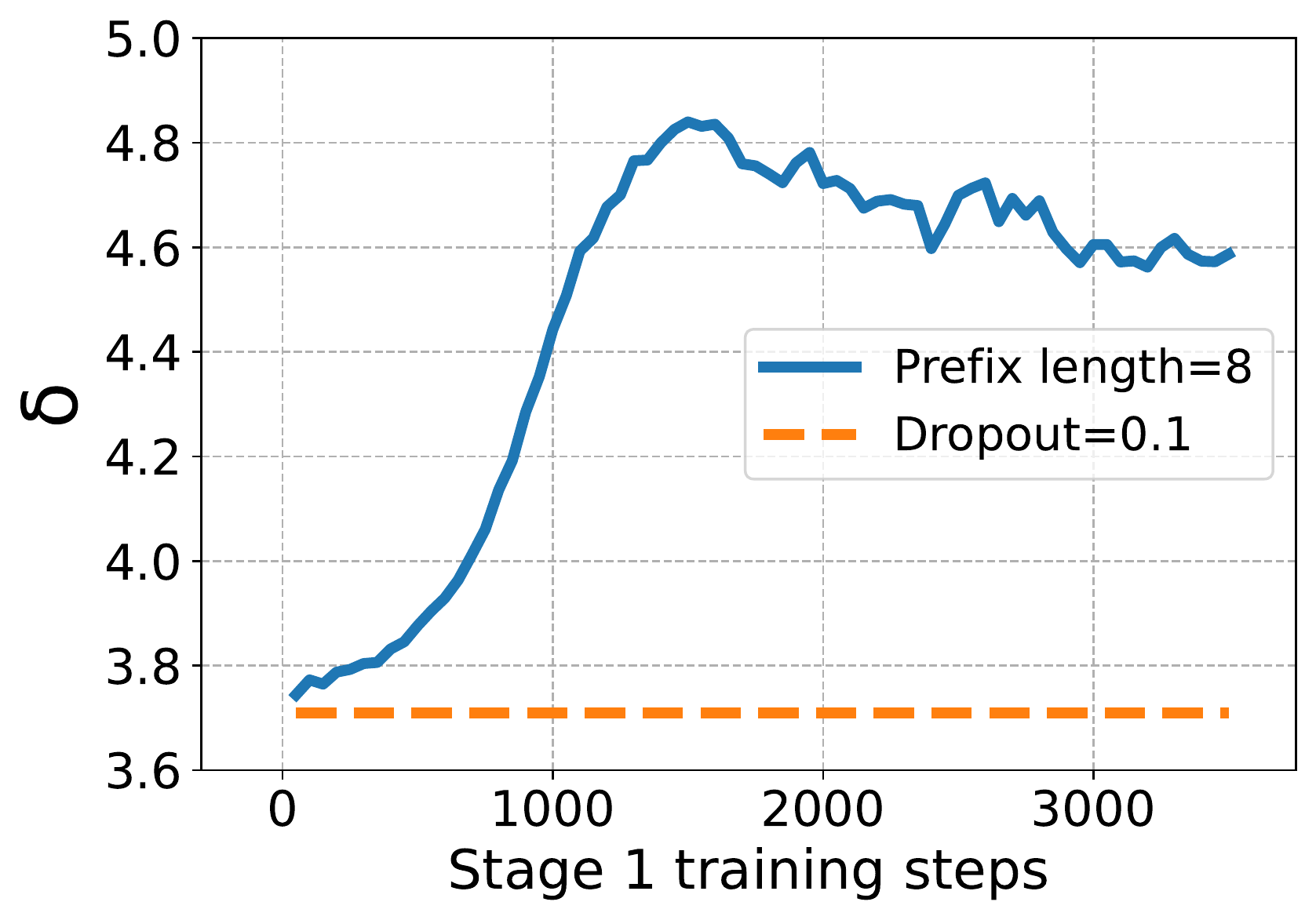}
      \captionsetup{font={scriptsize}}
      \caption{\label{fig:positive_difference} Representation divergence.}
    \end{subfigure}
    % \hspace{+0.3cm}
    \begin{subfigure}{.48\textwidth}
      \centering
      \includegraphics[width=\textwidth]{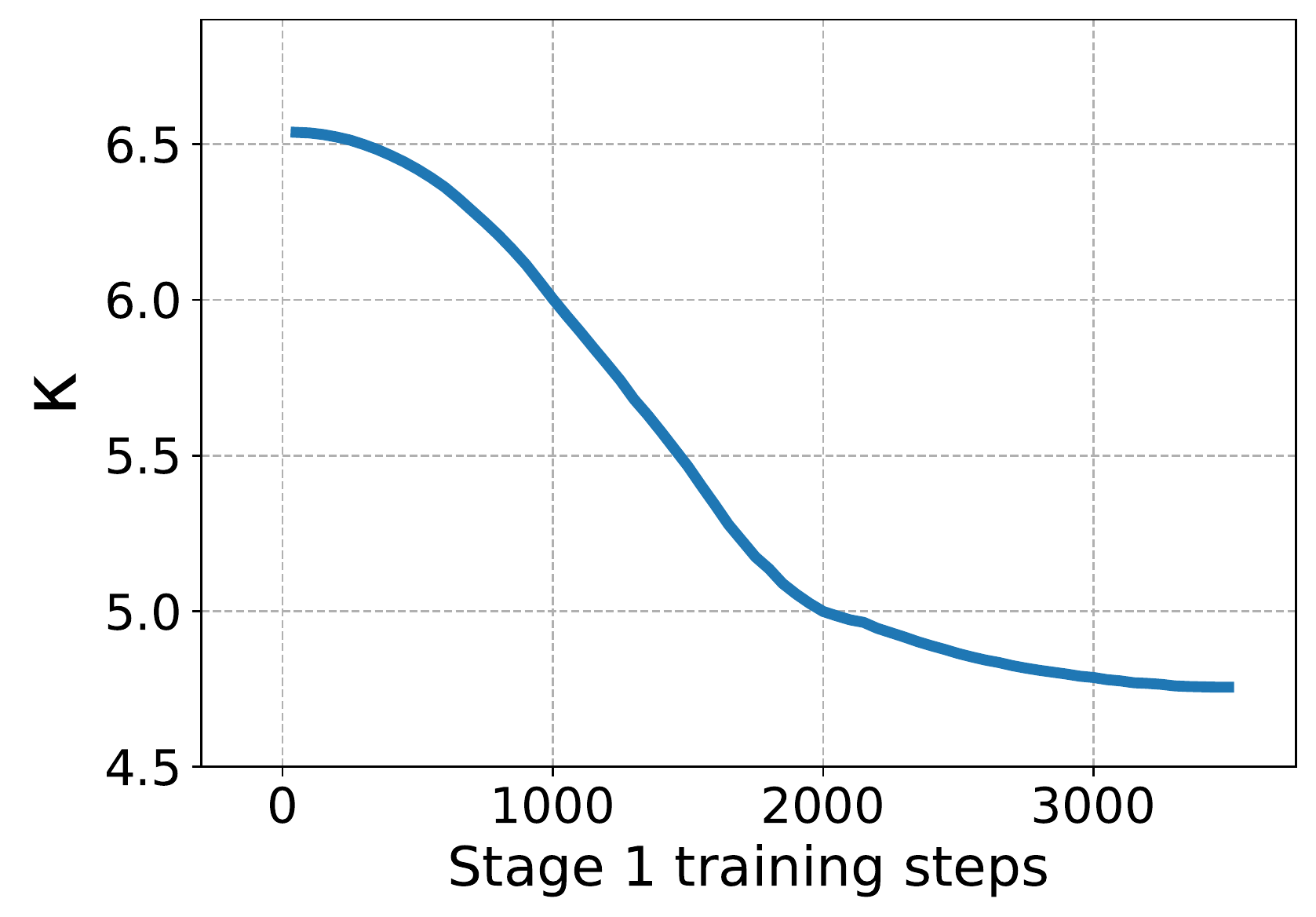}
      \captionsetup{font={scriptsize}}
      \caption{\label{fig:distribution_alignment} Weight convergence.}
    \end{subfigure}
    \caption{The visualization of the relationship between two metrics and stage-1 training steps from the BERT\ba \ with batch size 128 and learning rate 0.001.}
  \end{minipage}\hfill
\end{figure}

    Representation divergence means the difference of embeddings that are generated from a positive pair.
    Since the motivation of this work is to design a stronger data augmentation strategy than dropout~\cite{gao2021simcse}, a larger representation divergence is desired.
    To measure this metric quantitatively, we use the expected distance between representations generated from positive pairs:

\DeclarePairedDelimiterX{\infdivx}[2]{\Big(}{\Big)}{%
  #1\;\delimsize\|\;#2%
}
\newcommand{\infdiv}{\mathcal{D}\infdivx}
\DeclarePairedDelimiter{\norm}{\lVert}{\rVert}
\vspace{-10pt}
\begin{equation}
\delta \triangleq \mathop{\mathbb{E}}_{(x, x^+) \sim p_{\text{pos}}}
  ||f(x; \theta_1, \phi) - f(x^+; \theta_2, \phi)||_2   
\end{equation}
    We plot in Figure~\ref{fig:positive_difference} the trend of $\delta$ as stage 1 proceeds.
    As a reference, we also plot the representation divergence when dropout is used for data augmentation~\cite{gao2021simcse}.
    We can observe that $\delta$ quickly reaches the highest value at around 1,500 steps, and then its value slowly goes down.
    We performed multiple runs with different random seeds, and found the figures all suggest that the optimal number of steps would be around 1,500.
    However, 
    though we hope the representations of two positives could be as different as possible due to the InfoMin Principle~\cite{tian2020infomin}, we also found that in practice sometimes it would be beneficial to prolong the first stage with slightly more than 1,500 steps. 
    For example, the optimal number of stage-1 training steps for our semi-supervised model is 2,000.
    To explain this discrepancy, we now turn to look at the second metric, weight convergence.

    The prefix modules are initialized randomly,
    therefore it is unsurprising to find that the parameters of prefix modules are significantly different from those of the pre-trained language model.
    Specifically, both prefix and language model's key and value matrices have zero means, but those matrices of language model have a much larger variance than those of prefix.\footnote{
\citet{tao2022compression} made a similar observation about the distribution of weights (i.e., zero mean and large variance) in GPT-2 in Figure 4 of their paper.
}
We quantify this difference by measuring the weight convergence $\kappa$ -- the $l_{2,1}$-norm difference between the key-value matrices of prefix and those of language model. The weight convergence between $m$-th layer's key matrix of language model $K^m \in \mathbb{R}^{L \times d}$ and that of prefix $P^m_k \in \mathbb{R}^{l \times d}$ can be defined as follows:

% \vspace{-8pt}
\begin{equation}
\kappa^m_k \triangleq \frac{1}{L} ||K^m||_{2,1} - \frac{1}{l} ||P^m_k||_{2,1}
\end{equation}
where $||\cdot||_{2,1}$ represents the $l_{2,1}$-norm\footnote{For a matrix $A \in \mathbb{R}^{r \times c}$, its $l_{2,1}$-norm is defined as $||A||_{2,1} = \sum_{i=1}^r \sqrt{\sum_{j=1}^c (A_{ij})^2}$}.
The weight convergence between the value matrices $\kappa^m_v$ is defined similarly. In practice, the overall $\kappa$ is defined as the averaged score over the weight convergence for both key and value matrices on all $M$ self-attention layers:

\begin{equation}
\kappa \triangleq \frac{1}{2M} \sum_{m=1}^M \kappa_{k}^m + \kappa_{v}^m
\end{equation}

From Figure~\ref{fig:distribution_alignment}, we find the difference is monotonically decreasing, which indicates that the distribution gap reduces as stage-1 tuning continues.
Previous work~\cite{glorot2010init} has identified that such a parameter distribution gap can block the gradient descent based training, and we show how such distribution gap will affect the self-attention layer in details in Appendix~\ref{app:resonance}. 
Overall, the lower the $\kappa$, the easier it is for the model to perform stage-2 tuning. 
This explains why it could be worthwhile to start stage-2 tuning slightly after we have reached the peak for representation divergence -- though we have missed the peak, the prolonged stage-1 tuning that leads to an even lower $\kappa$ can potentially make the stage-2 tuning easier.

From the above analysis, we can see the module-compatible state is determined by both
$\delta$ and $\kappa$.
A larger $\delta$ will better explore the potential of the contrastive learning framework, while a smaller $\kappa$ will make the stage-2 tuning more stable.
An overall good performance is obtained when a module-compatible state is reached.
More details about the optimal stage-1 step are in Appendix~\ref{app:stage1steps}.

\section{Experiments}

    Following previous works,
    we mainly conduct experiments on Semantic Textual Similarity (STS) tasks.
    The results demonstrate the capability of our
    method on producing high-quality universal sentence representations.

\subsection{Experimental setup}

    We conduct experiments with both BERT\ba \ and RoBERTa\ba, but only present BERT\ba \  results in this section. The results of RoBERTa\ba \ can be found in Appendix~\ref{app:roberta}.
    We consider three different settings: unsupervised, semi-supervised, and supervised settings. 
    The unsupervised category contains both contrastive learning based and non-contrastive learning based methods that are trained with unlabeled examples.
    The semi-supervised setting allows the usage of both labeled and unlabeled data, but labeled data is only used for cross entropy loss.
    Our supervised setting only considers the supervised contrastive learning methods where the label information from NLI dataset is used for constructing positive and hard negative instances.

\paragraph{Datasets}

Following previous works,
we use two training datasets for the abovementioned settings: one is the unlabeled Wikipedia dataset (Wiki1M)~\cite{gao2021simcse} which contains $10^6$ sentences sampled from Wikipedia; the other is the labeled NLI dataset combining SNLI~\cite{bowman2015snli} and MNLI~\cite{williams2018mnli}. It comprises 275,600 sentence triplets with the format (anchor, entailment, contradiction).

We evaluate our method on seven standard STS datasets: STS tasks 2012-2016~\cite{agirre2012semeval,agirre2013semeval,agirre2014semeval,agirre2015semeval,agirre2016semeval}, STS-Benchmark~\cite{cer2017semeval}, and SICK-Relateness~\cite{marelli2014sick}. 
The sentence pairs in each dataset are scored from 0 to 5 to indicate the semantic similarity. Following previous works~\citep{gao2021simcse,jiang2022promptbert}, we report the Spearman's rank correlation coefficients between the cosine similarity of the learned sentence representations and the golden similarity scores on each dataset.

\begin{table*}[t]
    \setlength\tabcolsep{3pt}
    \begin{center}
    \centering
    \small
    \begin{tabular}{p{0.18\textwidth}cccccccc}
    \toprule
       Method & STS12 & STS13 & STS14 & STS15 & STS16 & STS-B & SICK-R & Avg. \\
    \midrule
        \multicolumn{9}{c}{\it{Unsupervised methods}}\\
    \midrule
        BERT$^\diamondsuit$  {\scriptsize(\texttt{[CLS]})}  
                        &  20.16&   30.01&  20.09&  36.88&  38.08&  16.50&  42.63&  29.19\\
        BERT$^\diamondsuit$  {\scriptsize(Mean Pooling)}      
                        &  38.78&   57.98&  57.98&  63.15&  61.06&  46.35&  58.40&  54.81\\
        BERT-flow$^\heartsuit$           &  58.40&   67.10&  60.85&  75.16&  71.22&  68.66&  64.47&  66.55 \\
        ConSERT$^\clubsuit$              &  66.07&   77.40&  70.32&  80.91&  77.29&  75.78&  66.91&  73.53 \\
        SimCSE                           &  68.40&   82.41&  74.38&  80.91&  78.56&  76.85&  72.23&  76.25 \\
        PromptBERT                       &  71.56&   84.58&  76.98&  84.47&  80.60&  81.60&  69.87&  78.54 \\
        DiffCSE         &  72.28&   84.43&  76.47&  83.90&  80.54&  80.59&  71.23&  78.49 \\
    \midrule
        \multicolumn{9}{c}{\it{Semi-supervised methods}}\\
    \midrule
        SBERT$^\heartsuit$               &  70.97&   76.53&  73.19&  79.09&  74.30&  77.03&  72.91&  74.89 \\
        SBERT-flow$^\heartsuit$          &  69.78&   77.27&  74.35&  82.01&  77.46&  79.12&  76.21&  76.60 \\
        ConSERT$_\text{joint}$$^\clubsuit$          
                                         &  72.60&   82.40&  77.64&  82.82&  78.37&  80.03&  \tf{78.20}&  78.87 \\
        DiffAug ({\em ours}) 
        &  \tf{75.28}\tiny$_{\pm.0}$&  85.20\tiny$_{\pm.2}$&  78.57\tiny$_{\pm.0}$
        &  85.20\tiny$_{\pm.2}$&   \tf{81.01}\tiny$_{\pm.3}$&  82.42\tiny$_{\pm.3}$&  73.24\tiny$_{\pm.5}$&  80.13\tiny$_{\pm.1}$ \\
        \ \ \ \ w/ aux loss          
        & 73.46\tiny$_{\pm.5}$& \tf{85.61}\tiny$_{\pm.2}$ & \tf{78.93}\tiny$_{\pm.3}$& \tf{85.82}\tiny$_{\pm.1}$& 80.74\tiny$_{\pm.2}$& 
        \tf{82.70}\tiny$_{\pm.2}$& 77.00\tiny$_{\pm.5}$& \tf{80.61}\tiny$_{\pm.2}$ \\
    \midrule
        \multicolumn{9}{c}{\it{Supervised contrastive learing methods}}\\
    \midrule
        SimCSE
        &75.30& 84.67& 80.19 & 85.40 & 80.82 &84.25 &80.39&81.57 \\
        PromptBERT
        &75.48& \tf{85.59}& 80.57 & 85.99 & 81.08 &\tf{84.56} &\tf{80.52}&81.97 \\
        DiffAug ({\em ours}) & \tf{76.92}\tiny$_{\pm.7}$& 85.17\tiny$_{\pm.2}$ & \tf{80.81}\tiny$_{\pm.4}$& \tf{86.91}\tiny$_{\pm.5}$& \tf{82.52}\tiny$_{\pm.1}$& 84.32\tiny$_{\pm.5}$& 80.27\tiny$_{\pm.2}$& \tf{82.42}\tiny$_{\pm.0}$ \\
    \bottomrule
    \end{tabular}
    \caption{
        \label{tab:main_sts}
        Sentence embedding performance on the seven standard STS tasks (Spearman's correlation).
        The highest results under semi-supervised and supervised settings are highlighted. 
        $^\diamondsuit$: results are from~\cite{reimers2019sbert};
        $^\heartsuit$: results are re-evaluated by~\cite{gao2021simcse}; 
        $^\clubsuit$: results are re-evaluated by us using the released code \protect\footnotemark. More details about the re-evaluation are in Appendix~\ref{app:semi-reimpl}.
        Other baseline results are from their original papers. All our results in this table are averaged over three different runs (with standard deviation).
    }
    \end{center}
\end{table*}

\paragraph{Baselines}
    We first consider two common methods for obtaining sentence embeddings from BERT: using \texttt{[CLS]} vector and mean pooling.
    \tf{BERT-flow}~\cite{li2020bertflow} is an unsupervised post-processing method that transforms the original BERT sentence embedding distribution to a smooth and isotropic Gaussian distribution.
    \tf{ConSERT}~\cite{yan2021consert} is an unsupervised contrastive learning method with several data augmentation techniques for sentences. Its unsupervised results can be further improved by incorporating NLI supervision.
    \tf{SimCSE}~\cite{gao2021simcse} is a contrastive learning framework that uses dropout for data augmentation. Its supervised approach regards the entailment sentence pairs from the NLI dataset as positives, and the contradiction pairs as hard negatives. 
    \tf{PromptBERT}~\cite{jiang2022promptbert} improves SimCSE by applying effective hard prompts on input sentences.
    \tf{DiffCSE}~\cite{chuang2022diffcse} combines the contrastive learning objective with a difference prediction objective which is inspired by equivariant contrastive learning~\cite{dangovski2021equivariant}.
    \tf{SBERT}~\cite{reimers2019sbert} uses BERT with a siamese structure to generate sentence embeddings after learning on NLI data with cross entropy loss. This method can be combined with the post-processing methods, e.g., BERT-flow~\cite{li2020bertflow}, to produce better results under the semi-supervised setting.

    \footnotetext{\url{https://github.com/yym6472/ConSERT}}

\subsection{Main results}

    We compare our model with previous sentence embedding methods on the standard seven STS datasets in Table~\ref{tab:main_sts}.
    Since our method requires a supervised prefix-tuning stage before the contrastive learning, 
    we only report our results under the semi-supervised and supervised settings. 
    For semi-supervised setting, we also report the results of our method with the auxiliary loss that is mentioned in Section~\ref{sec:stage2}.
    From Table~\ref{tab:main_sts}, we make the following observations:

    \begin{itemize}[topsep=0pt, partopsep=0pt, leftmargin=10pt, parsep=0pt, itemsep=5pt]

        \item The quality of sentence embeddings directly obtained from BERT\ba \ (both \texttt{[CLS]} vector and mean pooling) without further fine-tuning is poor. This phenomenon and the reasons behind it have been studied extensively~\cite{reimers2019sbert,li2020bertflow}.

        \item In general, supervised and semi-supervised models are better than unsupervised models,
        which indicates that the label information in NLI data is beneficial to learning good sentence embeddings. 
        
        \item Our method outperforms baselines in both semi-supervised and supervised settings on average. Since our method strictly follows the contrastive learning framework proposed by~\citet{gao2021simcse} except for the data augmentation module, our improved results confirm that the proposed data augmentation method is effective for contrastive learning.

    \end{itemize}

\subsection{Training with less labeled data} \label{sec:low_data}

    In this section, we show our method is more label-efficient than the state-of-the-art contrastive learning methods with BERT\ba. The results are presented in Table~\ref{tab:low_data}. 
    We find not all the previous contrastive learning methods discuss the semi-supervised setting. 
    However, \citet{yan2021consert} proposes several ways of adding supervised signal into their unsupervised contrastive learning, and these methods are also applicable to other unsupervised approaches. 
    Therefore, we re-evaluate previous methods under the semi-supervised setting following~\cite{yan2021consert}, and produce stronger baselines. The details about the re-implementation can be found in Appendix~\ref{app:semi-reimpl}. 
    We report the results of our semi-supervised method with auxiliary loss in stage 2. 
    For previous supervised contrastive learning methods that do not discuss the low labeled data condition, we re-evaluate their methods with less labeled data. 
    
    To construct the low labeled data settings, we sub-sample 1\% ($\sim$2,756) or 10\% ($\sim$27,560) labeled examples from the full NLI dataset.
    For each size, we sample 5 sub-datasets and take the average over 3 different runs. 
    Thus, we report the averaged results over 15 models under each low-data setting.
    Table~\ref{tab:low_data} shows our approach significantly outperforms baselines under all settings.
    We also notice that the performance gap increases as the number of labeled examples decreases.
    
    We suspect that the advantages of our proposed method in low labeled data situations come from the prefix-tuning in stage 1.
    Since the baseline methods all fine-tune the whole parameters of the language model, it is easy to overfit when the labeled data is scarce. 
    However, for our method, the amount of added prefix parameters is much smaller (only 0.3\% of the parameters in BERT\ba).
    Therefore, the overfitting problem can be alleviated.

\subsection{Ablation study}

    In this section, we compare several variants of our proposed prefix-based method, and
    investigate the impact of different parameter-efficient methods for data augmentation and prefix length.
    More ablation studies (auxiliary loss and stage-1 training steps) are provided in Appendix~\ref{app:abl}.
    All results are based on the STS-B development set.

\begin{table}[!t]
    \setlength\tabcolsep{3pt}
    \small
    \centering
    \begin{tabular}{lcccc}
        \toprule
          \multirow{2}{*}{Method}   & \multicolumn{4}{c}{Fraction of labeled data}\\
          \cmidrule{2-5}
                                    & 0\%                   & 1\%                       & 10\%                      & 100\%\\
        \midrule
          \multicolumn{5}{c}{\textit{Semi-supervised setting}}\\
        \midrule
          SimCSE$_\text{joint}$     & 76.25$^\heartsuit$    & 76.45\tiny$_{\pm.6}$      & 77.46\tiny$_{\pm.6}$      & 77.73\tiny$_{\pm.6}$ \\
          PromptBERT$_\text{joint}$ & 78.54$^\spadesuit$    & 79.26\tiny$_{\pm.3}$      & 79.65\tiny$_{\pm.3}$      & 79.81\tiny$_{\pm.1}$ \\
          DiffAug ({\em ours})      & -                     & \tf{80.42}\tiny$_{\pm.1}$ & \tf{80.58}\tiny$_{\pm.1}$ & \tf{80.61}\tiny$_{\pm.2}$\\
        \midrule
          \multicolumn{5}{c}{\textit{Supervised setting}}\\
        \midrule
          SimCSE                    & -                     & 75.26\tiny$_{\pm.4}$      & 79.04\tiny$_{\pm.2}$      & 81.58\tiny$_{\pm.1}$\\ 
          PromptBERT                & -                     & 78.33\tiny$_{\pm.2}$      & 80.01\tiny$_{\pm.1}$      & 81.88\tiny$_{\pm.1}$\\
          DiffAug ({\em ours})      & -                     & \tf{80.18}\tiny$_{\pm.6}$ & \tf{81.73}\tiny$_{\pm.2}$ & \tf{82.42}\tiny$_{\pm.0}$\\
        \bottomrule
    \end{tabular}
    \caption{
        \label{tab:low_data} 
        Averaged sentence representation performance on the standard seven STS tasks with different sizes of labeled data. 
        $^\heartsuit$ and $^\spadesuit$ are the results of unsupervised SimCSE and PromptBERT respectively. 
        We report these two results from their original papers. All other baseline results are re-evaluated by us. 
    }
\end{table}

    \paragraph{Variants of applying prefix.}
    In this work, we propose to use prefix-tuning~\cite{li2021prefix} for data augmentation.
    Now we investigate the performance of several variants of the proposed method.
    The results are presented in Table~\ref{tab:data_aug}. 
    The amount of added parameters relative to that in BERT\ba \ is also reported. 
    We select two unsupervised data augmentation methods, i.e., synonym replacement and dropout, as baselines since they are simple yet effective for sentence representation learning.
    
    Our method trains two {\em different} prefix modules in stage 1 and {\em jointly} tunes both prefixes and language model in stage 2. 
    Here we first consider two variants of applying prefixes: 
    using two identical prefixes (same) and fixing the prefixes during stage 2 (fix). 
    The performance gap between the first variant and our method confirms that contrastive learning performs better when the positives are meaningfully different. 
    The performance drop caused by the second variant validates the necessity of tuning prefix and language model together in stage 2.

    To investigate the importance of fine-tuning language model with prefix in stage 2, 
    we consider another prefix variant, i.e., prefix-tuning, which means that only prefix modules will be tuned for both stage 1 and stage 2. 
    The performance gap between our proposed method and the ``prefix-tuning'' method indicates that optimizing the language model is necessary for obtaining good results on sentence representation learning tasks.

    \paragraph{Different parameter-efficient methods as data augmentation module.} \label{sec:abl-pe}

    We also consider other parameter-efficient (PE) methods for data augmentation. 
    Previous work~\cite{he2021padapter} has unified current PE methods under one framework. 
    Hence, it is theoretically possible to transfer the success of our method with prefix-tuning to other PE methods. 
    In this section, we additionally tried two most common and effective methods, i.e., Adapter~\cite{houlsby2019adapter} and LoRA~\cite{hu2021lora} with different bottleneck dimensions.
    The description of these two methods can be found in Appendix~\ref{app:PE}.

    As shown in Table~\ref{tab:data_aug}, 
    with a small amount of additional parameters ($0.3\%$), data augmentation with distinct prefix modules obtains the highest score on the STS-B development set. 
    Such a phenomena of prefix's effectiveness with small budget of additional parameters has also been observed by both~\citet{he2021padapter} and~\citet{li2021prefix} on textual generation tasks. 
    We believe this is because prefix directly modifies each attention head on the multi-head attention layers, which makes it more expressive than other methods when parameter budget is limited.
    
    We then study whether this conclusion changes if more tunable parameters are added.
    Table~\ref{tab:data_aug} shows that the performance of Adapter drops after having more tunable parameters ($0.3\% \rightarrow 4.8\%$), but LoRA performs better (STS-B dev. score improves by $0.7$ points) when the bottleneck dimension increases from $4$ to $64$, though it is still slightly lower than our proposed prefix-based method.

    From the above results, we conclude that applying prefixes for data augmentation is the most effective and space-efficient method for sentence representation learning tasks.

\begin{table}[t]
  \setlength\tabcolsep{4pt}
  \centering
  \centering
  \small

  \begin{tabular}{lcc}
  \toprule
  \multirow{2}{*}{Method}   & \# added params & \multirow{2}{*}{STS-B dev.}\\
                                           & (rel. BERT\ba)  &   \\
  \midrule
    \multicolumn{3}{l}{\textit{Unsupervised methods}}\\
    Synonym replacement$^\heartsuit$       & 0       &  77.4  \\
    Dropout$^\heartsuit$                   & 0       &  82.5  \\
  \midrule
    \multicolumn{3}{l}{\textit{Prefix variants ($l=8$)}}\\
    Prefix (same)                   & 0.3\%   &    85.7  \\
    Prefix (fix)                    & 0.3\%   &    85.4  \\
    Prefix (different, {\em ours})  & 0.3\%   & \tf{86.2} \\
    Prefix-tuning (different)       & 0.3\%   &    84.8  \\
  \midrule
    \multicolumn{3}{l}{\textit{Other parameter-efficient methods}}\\
    Adapter (different, $r=4$)             & 0.3\%   &    85.6  \\
    Adapter (different, $r=64$)            & 4.8\%   &    85.1  \\
    LoRA    (different, $r=4$)             & 0.3\%   &    85.4  \\
    LoRA    (different, $r=64$)            & 4.8\%   &    86.1  \\
  \bottomrule
  \end{tabular}
  \caption{\label{tab:data_aug} 
  Comparison between different methods for data augmentation under the semi-supervised setting. 
  $l$ and $r$ are the prefix length and bottleneck dimension respectively.
  {\small$^\heartsuit$}: results are from \citet{gao2021simcse}.}
\end{table}

\paragraph{Prefix length.}

\begin{figure}[t]
    \captionsetup{type=figure}
    \centering
    \includegraphics[width=.80\linewidth]{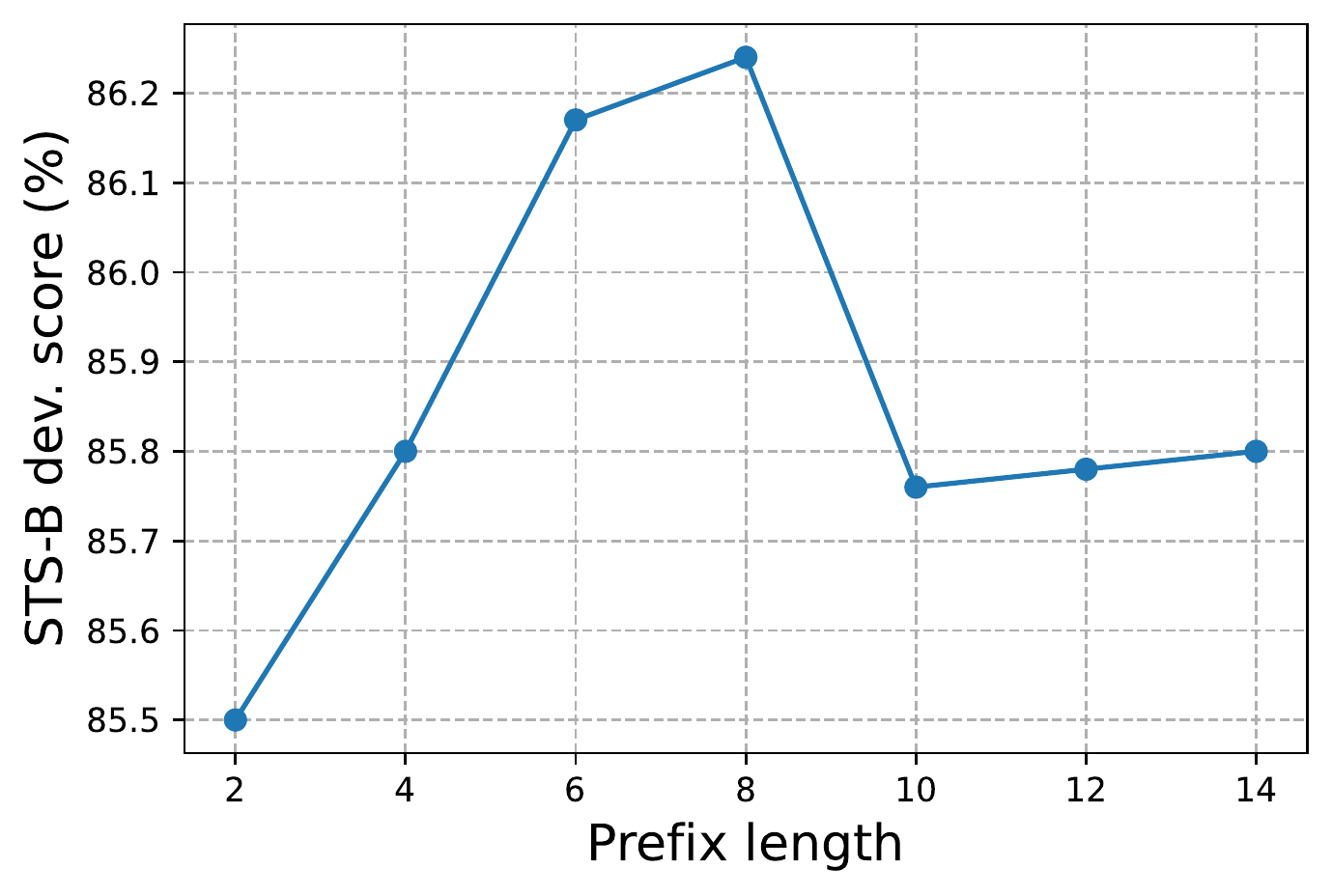}
    \caption{\label{fig:prefix_len} The relationship between prefix length and STS-B development set performance under the semi-supervised setting using BERT\ba.}
\end{figure}

Prefix length is a critical hyperparameter in our experiments. The longer the prefix, the more tunable parameters, therefore the more expressive the data augmentation module is. However, it does not mean a longer prefix will definitely lead to a better performance. According to the InfoMin principle~\cite{tian2020infomin}, a longer prefix may bring unnecessary noise, thus hurting the performance of contrastive learning. Figure~\ref{fig:prefix_len} shows that the optimal prefix length for BERT\ba \ under the semi-supervised setting is around 8.

\section{Related work}

    Learning sentence embeddings as a fundamental NLP problem has been extensively studied. 
    Many works~\cite{reimers2019sbert,li2020bertflow,yan2021consert,gao2021simcse,jiang2022promptbert} achieve good results based on pre-trained language models. Among these works, contrastive learning methods achieve the state-of-the-art results. In this work, we improve the data augmentation module of the contrastive learning framework by a prompting method.

    \paragraph{Contrastive sentence representation learning.}
    Contrastive learning aims to learn effective representations by pulling together semantically close neighbors and pushing apart non-neighbors. 
    A crucial problem in contrastive learning is how to generate positive instances. 
    The commonly-used data augmentation methods in NLP include word deletion, reordering, and substitution~\cite{xie2020uda,yan2021consert}. 
    However, data augmentation in NLP is inherently difficult because of the discrete nature of language. 
    \cite{gao2021simcse} shows that data augmentation via dropout is consistently better than previous discrete transformations. 

\paragraph{Prompting.}
    Prompting means querying a pre-trained language model by adding natural language tokens into the input sentences~\cite{brown2020gpt3}. 
    This method alleviates the discrepancy between a language model's pre-training stage and fine-tuning stage, and is shown to be useful when the labeled data for downstream tasks is scarce~\cite{gao2021lmbff}. 
    Although adding natural language tokens (i.e., hard prompt) is effective for certain tasks, it takes efforts to design good prompts.
    Motivated by this problem, soft prompt methods~\cite{qin2021learning,zhong2021optiprompt} are proposed. Unlike hard prompt, soft prompt tokens can be fine-tuned continuously. Therefore, it is more expressive.
    Some recent works further improve the soft prompt method by adding deep prompt modules on language models~\cite{houlsby2019adapter,li2021prefix,liu2021pt2}, and they are called parameter-efficient methods. 
    With more tunable parameters, the parameter-efficient methods can handle complex natural language tasks, e.g., language understanding~\cite{houlsby2019adapter}, text generation~\cite{li2021prefix}, and structured prediction~\cite{liu2021pt2}. 
    Such methods usually assume the language model is fixed during fine-tuning, and only the newly-added prompting modules can be tuned.

\section{Conclusion}
\label{sec:conclusion}

    In this paper, we propose a differentiable data augmentation method for contrastive sentence representation learning. 
    Unlike previous discrete transformations for sentences (e.g., token shuffling and synonyms replacement) and continuous methods (e.g., dropout~\cite{gao2021simcse}), the proposed method learns how to do data augmentation implicitly based on supervised signals from the natural language inference (NLI) tasks. 
    In this way, our method produces \textit{hard positives} with meaningful differences for contrastive learning. We demonstrate the effectiveness of our method on several semantic textual similarity tasks, and our method achieves new state-of-the-art performance in both semi-supervised and supervised settings on average.
    We also conduct experiments where only limited labeled data is available. The results demonstrate the proposed method is robust when is labeled data is scarce.

    To the best of our knowledge, our method is the first work that provides a successful solution for combining the conventional pre-trained language model fine-tuning with parameter-efficient methods via a two-stage tuning process.
    The approach of combining the two yields better results than using any of them alone on the sentence representation learning tasks.
    One potential future research direction is to generalize this method to other NLP research areas, e.g., domain adaptation and low-resource tasks.

\section*{Limitations}

Since our method incorporates a learnable module for data augmentation, it requires additional training time and GPU memory compared with previous contrastive learning methods. The demand for more computation resources is a clear limitation of our work. We compare the training overheads with two previous contrastive learning methods in Table~\ref{tab:limit}. We have mentioned in the previous section that our stage-1 tuning is necessary, and it is not a trivial task to reach a module-compatible state. Therefore, reducing computation resources during training for our method is difficult. However, given the improvements that are brought by our methods on sentence embedding learning tasks, the additional computation cost is worthy.
\begin{table}[h]
  \setlength\tabcolsep{5pt}
  \centering
  \small
  \begin{tabular}{lcc}
  \toprule
    Method                & Training time & GPU memory usage\\
  \midrule
    SimCSE                &  36 min.      & 10 GB \\
    PromptBERT            &  48 min.      & 13 GB \\
    DiffAug ({\em ours})  &  65 min.      & 15 GB \\
  \bottomrule
  \end{tabular}
  \caption{\label{tab:limit} Comparison of training time and GPU memory usage of our method and previous contrastive learning methods. We use same batch size (128) and same number of epochs (3) on NLI dataset with BERT\ba \ for all three methods for a fair comparison. All results in this table are obtained from a single Nvidia Quadro RTX8000 GPU.}
\end{table}

\section*{Acknowledgements}
We would like to thank the anonymous reviewers and (senior) area chairs for their insightful comments and help with this work.
We would also like to thank members of StatNLP research group for helpful discussions.
This research is supported by the National Research Foundation, Singapore under its AI Singapore Programme (AISG Award No: AISGRP-2019-012). Any opinions, findings and conclusions or recommendations expressed in this material are those of the authors and do not reflect the views of National Research Foundation, Singapore and AI Singapore.

\bibliography{anthology,custom}
\bibliographystyle{acl_natbib}

\appendix

% \clearpage
\section{Implementation details}
\label{app:impl_details}

We implement our method based on Hugging Face's Transformer~\cite{wolf2020hf} and PyTorch libraries. 
The backbone of our code is from~\cite{gao2021simcse} with some prompt implementation details from~\cite{jiang2022promptbert}. 
For both semi-supervised and supervised settings, we evaluate the model on the STS-B development set every 50 training steps, and only save the checkpoints with the best performance on development set for the final evaluation on test set.

For all our settings, we apply hard prompts on the input sentences and obtain the sentence representations from the hidden embeddings of \mask \ tokens. 
The hard prompt templates we use in our method are from~\cite{jiang2022promptbert}. 
We also apply the template denoising for contrastive learning since this technique brings significant improvements from only adding hard prompts according to~\cite{jiang2022promptbert}. 

The core component of our method is the use of two different prefix modules for data augmentation. 
During training, we use them to construct hard positive pairs with meaningful differences for contrastive learning.
During inference, we prepend both prefix modules to the languge model to obtain the sentence representations.
In practice, we follow~\citet{li2021prefix} and reparametrize the key and value matrices $P_k$, $P_v$ in the prefix by a smaller matrix $P'$ composed with a large multi-layer perceptron to make the training stable.
Our training hyperparameters under different settings for BERT\ba \ are in Table~\ref{tab:hyperpara}.

    \begin{table}[h]
    \small
    \centering
    \begin{tabular}{clccc}
    \toprule
      & \multirow{2}{*}{}        & \multicolumn{2}{c}{\tf{Semi-supervised}} & \tf{Supervised} \\
      &                          & w/o aux            & w/ aux              & w/o aux         \\
    \midrule
    \multirow{3}{*}{\rotatebox[origin=c]{90}{\small Stage 1}}
      & \small learning rate              & 1e-3           & 1e-3              & 1e-3    \\
      & \small batch size                 & 128            & 128               & 128     \\
      & \small training steps             & 2000           & 1500              & 1250    \\
    \midrule
    \multirow{3}{*}{\rotatebox[origin=c]{90}{\small Stage 2}}
      & \small learning rate              & 1e-5            & 1e-5            & 5e-5      \\
      & \small batch size                 & 256             & 256             & 128       \\
      & \small epochs                     & 1               & 1               & 3         \\
    \bottomrule
    \end{tabular}
    \caption{\label{tab:hyperpara} Hyperparameters for each setting.}
    \end{table}

\section{Analysis of weight convergence}
\label{app:resonance}

In the previous section, we mentioned that the module-compatible state in our case is closely related to the stage-1 training steps, and the optimal number of training steps is determined by two countering metrics, i.e., representation divergence and weight convergence. Previous works~\cite{chen2020simclrv1,tian2020infomin} have demonstrated the importance of representation divergence in contrastive learning. In this section, we show the necessity of weight convergence in our stage-2 training. Our method is based on the prefix-tuning~\cite{li2021prefix} that affects the original language model by adding new key-value pairs on each self-attention layer. The added key and value matrices influence the self-attention computation differently. Specifically, key matrix ($K$) is used for attention matrix calculation, i.e., $A = \softmax(\frac{Q^T K}{\sqrt{d}})$ while value matrix ($V$) is used for weighted averaged output generation, i.e., $H = AV$, we analyze the impact of weight convergence on key and value separately.

\subsection{The impact of weight convergence on attention matrix calculation}

The computation of attention matrix includes a softmax operation $S(\cdot): \mathbb{R}^N \rightarrow \mathbb{R}^N$ that can be defined as:

\begin{equation}
S(\mathbf{a}) :
\left(            
  \begin{array}{c}
    a_1 \\ \vdots \\ a_N
  \end{array}
\right)
\rightarrow
\left(            
  \begin{array}{c}
    s_1 \\ \vdots \\ s_N
  \end{array}
\right)
\end{equation}

\noindent
where $s_i = \frac{\exp(a_i)}{\sum_{j=1}^{N}\exp(a_j)}$, and $N$ is the length of input and output vectors. Therefore, the derivative of the softmax function $S(\cdot)$, i.e., the Jacobian matrix, can be derived as:
\begin{equation}
J_{\softmax} =
\left(            
  \begin{array}{cccc}
    \frac{\partial s_1}{\partial a_1} & \frac{\partial s_1}{\partial a_2} & \hdots & \frac{\partial s_1}{\partial a_N}\\
    \frac{\partial s_2}{\partial a_1} & \frac{\partial s_2}{\partial a_2} & \hdots & \frac{\partial s_2}{\partial a_N}\\ 
    \vdots & \vdots & \ddots & \vdots \\ 
    \frac{\partial s_N}{\partial a_1} & \frac{\partial s_N}{\partial a_2} & \hdots & \frac{\partial s_N}{\partial a_N}
  \end{array}
\right)
\end{equation}

\noindent
where
\begin{equation}
\frac{\partial s_i}{\partial a_j} = 
\left\{
  \begin{array}{r}
    s_i \cdot (1-s_j), \text{if } i=j \\
    - s_i \cdot s_j, \text{if } i \neq j \\
  \end{array}
\right.
\end{equation}

From the above derivation, either a too large or a too small $s_i$ will make the derivative approach zero. Notice that in Section~\ref{sec:why-work}, we mentioned the averaged norm of the key vectors from randomly initialized prefix module is much smaller than those in the original language model. Therefore, if we tune both prefix and language model jointly without stage-1 training, the gradient back propagation of the self-attention layer will be blocked due to the vanishing gradients.

\begin{table*}[t]
    \setlength\tabcolsep{3pt}
    \begin{center}
    \centering
    \small
    \begin{tabular}{p{0.25\textwidth}cccccccc}
    \toprule
       Method                   &  STS12& STS13 & STS14 & STS15 & STS16 & STS-B & SICK-R& Avg. \\
    
    \midrule
        ConSERT$_\text{joint}$  &  \tf{72.60}$_{\pm.4}$&  82.40$_{\pm.8}$&  \tf{77.64}$_{\pm.5}$
                                &       82.82$_{\pm.1}$&  \tf{78.37}$_{\pm.1}$&  80.03$_{\pm.4}$
                                &  \tf{78.20}$_{\pm.1}$&  \tf{78.87}$_{\pm.3}$\\
        
        ConSERT$_\text{sup-unsup}$     &70.11\tiny$_{\pm.2}$&82.07\tiny$_{\pm.5}$&74.97\tiny$_{\pm.4}$
                                &  \tf{83.21}\tiny$_{\pm.2}$&  76.91\tiny$_{\pm.3}$&  78.89\tiny$_{\pm.3}$
                                &  76.92\tiny$_{\pm.5}$&  77.58\tiny$_{\pm.3}$\\
        
        ConSERT$_\text{sup-joint}$     &71.96\tiny$_{\pm.3}$&\tf{82.70}\tiny$_{\pm.4}$&75.85\tiny$_{\pm1.1}$
                                &82.70\tiny$_{\pm.5}$&77.16\tiny$_{\pm.8}$&  \tf{80.78}\tiny$_{\pm.7}$&75.12\tiny$_{\pm.4}$&78.04\tiny$_{\pm.5}$\\
    
    \midrule
        SimCSE$_\text{joint}$   &  70.73\tiny$_{\pm1.8}$& \tf{83.22}\tiny$_{\pm.3}$&  \tf{75.75}\tiny$_{\pm.5}$
                                &  \tf{82.93}\tiny$_{\pm.6}$&  \tf{77.90}\tiny$_{\pm1.0}$&  \tf{78.69}\tiny$_{\pm.3}$
                                &  \tf{74.86}\tiny$_{\pm.5}$&  \tf{77.73}\tiny$_{\pm.6}$\\
        
        SimCSE$_\text{sup-unsup}$      &  69.02\tiny$_{\pm2.4}$&  82.21\tiny$_{\pm1.3}$&  74.51\tiny$_{\pm.9}$
                                &  82.27\tiny$_{\pm.6}$&  77.20\tiny$_{\pm.6}$&  78.55\tiny$_{\pm1.3}$
                                &  74.06\tiny$_{\pm1.8}$&  76.83\tiny$_{\pm1.3}$\\
        
        SimCSE$_\text{sup-joint}$      &  \tf{70.80}\tiny$_{\pm1.2}$&  81.27\tiny$_{\pm.2}$&  74.68\tiny$_{\pm.5}$
                                &  81.83\tiny$_{\pm1.0}$&  77.58\tiny$_{\pm1.2}$&  78.55\tiny$_{\pm1.5}$
                                &  73.58\tiny$_{\pm1.1}$&  76.90\tiny$_{\pm.8}$\\
    \midrule
        PromptBERT$_\text{joint}$      &  \tf{71.71}\tiny$_{\pm.3}$&  \tf{84.99}\tiny$_{\pm.1}$&  \tf{77.25}\tiny$_{\pm.2}$
                                &  \tf{85.01}\tiny$_{\pm.1}$&  \tf{80.66}\tiny$_{\pm.1}$&  \tf{81.86}\tiny$_{\pm.1}$
                                &  \tf{75.50}\tiny$_{\pm.3}$&  \tf{79.81}\tiny$_{\pm.1}$\\
        
        PromptBERT$_\text{sup-unsup}$  &  71.23\tiny$_{\pm.2}$&  85.16\tiny$_{\pm.1}$&  77.88\tiny$_{\pm.2}$
                                &  85.15\tiny$_{\pm.2}$&  81.23\tiny$_{\pm.2}$&  82.06\tiny$_{\pm.6}$
                                &  75.22\tiny$_{\pm.8}$&  78.15\tiny$_{\pm.2}$\\
        
        PromptBERT$_\text{sup-joint}$  &  71.34\tiny$_{\pm.7}$&  81.61\tiny$_{\pm1.0}$&  76.23\tiny$_{\pm.6}$
                                &  82.92\tiny$_{\pm.5}$&  78.45\tiny$_{\pm.7}$&  79.76\tiny$_{\pm1.0}$
                                &  74.63\tiny$_{\pm.7}$&  77.85\tiny$_{\pm.7}$\\
    \bottomrule
    \end{tabular}
    \caption{
        \label{tab:semi-baselines}
        Semi-supervised baselines re-implemented by us over three different settings.
    }
    \end{center}
\end{table*}

\subsection{The impact of weight convergence on weighted averaged output generation}

In the self-attention layer of the Transformer, the embeddings of each token are generated via the weighted average of the value vectors. Previous works~\cite{glorot2010init,he2015delving} have shown the importance of keeping the same the variance of input of each layer in a deep neural network during forward propagation. Specifically, if we write the input matrix of layer $i$ as $Z^i$, to ensure the information flow, we hope 
\begin{equation} \label{eq:v}
\forall(i, i'), \ \var [Z^i] = \var[Z^{i'}]
\end{equation}

% \noindent
For the Transformer's $i$-th self-attention layer, we use $X^i, \ H^i \in \mathbb{R}^{L \times d}$ to represent the input and output respectively, where $L$ is the input length, and $d$ is the hidden dimension. 
The importance of weight convergence in attention matrix has been discussed in the last section. Now, for simplicity, we write the attention function as $H^i = A^i  [P_v^i;\ V^i]$, where $A^i \in \mathbb{R}^{L \times (L+l)}$ is the attention matrix after softmax, $P_v^i \in \mathbb{R}^{l\times d}$ is the added value matrix from the prefix module, and $V^i\in \mathbb{R}^{L\times d}$\footnote{$V^i  = X^i W_v^i$, where $W_v^i\in \mathbb{R}^{d\times d}$ is the projection matrix of value.} is the value matrix of the original language model.

We have shown in Section~\ref{sec:why-work} the norm of $P_v^i$ is significantly less than that of $V^i$ at the beginning. Therefore, if the training steps of stage 1 is not long enough, the difference between the variance of $P_v^i$ and $V^i$ could be so large that makes the condition in the Equation~\ref{eq:v} invalid, which will cause the stage-2 joint training unstable.

\section{Baseline methods}
\label{app:semi-reimpl}
In this section, we describe how we re-evaluate the baseline models when their original published results are not directly comparable with our methods. For unsupervised ConSERT~\cite{yan2021consert}, we re-evaluate it on Wiki1M dataset.

For methods that do not implement with semi-supervised setting, we find that simply adding an auxiliary cross entropy loss with proper weight can boost the performance from unsupervised results. To make a comprehensive evaluation, we try the following three methods of adding supervised signals from NLI data inspired by~\cite{yan2021consert}:
\begin{itemize}[topsep=0pt, partopsep=0pt, leftmargin=15pt, parsep=0pt, itemsep=5pt]

    \item \tf{Joint training (joint).} We combine the contrastive learning objective with the cross entropy loss, which can be written as
    \begin{equation}
      \mathcal{L}_{\text{joint}} = \mathcal{L}_{\text{cl}} + \alpha \cdot \mathcal{L}_{\text{ce}}
    \end{equation}
    where the contrastive loss is computed with unlabeled Wiki1M data while the cross entropy loss is computed with labeled NLI data. For this setting, we try different values of $\alpha$ for each method and select the best one to compare with our method.

    \item \tf{Supervised training then unsupervised contrastive training (sup-unsup).} We split the whole training process into 2 stages: first train the model with cross entropy loss on NLI data, then train the model with contrastive loss on Wiki1M data.
    
    \item \tf{Supervised training then joint training (sup-joint).} The training process is still split into 2 stages: first, the model is trained with cross entropy loss on NLI data; next, the model is trained with the joint loss $\mathcal{L}_{\text{joint}}$.

\end{itemize}
The results of each setting are shown in Table~\ref{tab:semi-baselines}. We select the setting with the best performance (i.e., joint training setting) and use it to compare with our method in Table~\ref{tab:low_data}.

\section{RoBERTa results} \label{app:roberta}
We compare our approach on RoBERTa\ba \ with the state-of-the-art contrastive learning methods~\cite{gao2021simcse,jiang2022promptbert} under different fractions of labeled data in Table~\ref{tab:low_data_r}.
We use a similar approach to evaluate our method and baselines on RoBERTa\ba \ that we described in Section~\ref{sec:low_data}.
The results show that the proposed method also outperforms the baselines on RoBERTa\ba.

\begin{table}[!t]
    \setlength\tabcolsep{3pt}
    \small
    \centering
    \begin{tabular}{lcccc}
        \toprule
      \multirow{2}{*}{Method}  & \multicolumn{4}{c}{Fraction of labeled data}\\
      \cmidrule{2-5}
                        & 0\%         & 1\%                & 10\%                       & 100\% \\
    \midrule
      \multicolumn{5}{c}{\textit{Semi-supervised setting}}\\
    \midrule
      SimCSE$_\text{joint}$                   & 76.57$^\heartsuit$ & 77.32\tiny$_{\pm.8}$ & 77.66\tiny$_{\pm.5}$      & 77.85\tiny$_{\pm.5}$ \\ 
      PromptRoBERTa$_\text{joint}$                & 79.15$^\spadesuit$ & 79.89\tiny$_{\pm.5}$ & 80.12\tiny$_{\pm.5}$      & 80.45\tiny$_{\pm.5}$ \\
      DiffAug ({\em ours})      & -                  & \tf{80.56}\tiny$_{\pm.3}$ & \tf{80.71}\tiny$_{\pm.3}$ & \tf{80.93}\tiny$_{\pm.3}$ \\
    \midrule
      \multicolumn{5}{c}{\textit{Supervised setting}}\\
    \midrule
      SimCSE                    & -          & 76.83\tiny$_{\pm.6}$  & 80.18\tiny$_{\pm.2}$             & 82.43\tiny$_{\pm.2}$\\ 
      PromptRoBERTa                & -          & 79.15\tiny$_{\pm.4}$  & 81.22\tiny$_{\pm.3}$             & 82.50\tiny$_{\pm.2}$\\ 
      DiffAug ({\em ours})      & -          & \tf{79.50}\tiny$_{\pm.7}$ & \tf{81.89}\tiny$_{\pm.2}$    & \tf{82.76}\tiny$_{\pm.2}$ \\
    \bottomrule
    \end{tabular}
    \caption{\label{tab:low_data_r} 
    Averaged sentence representation performance on STS tasks with different fractions of labeled data using RoBERTa\ba.
    $^\heartsuit$ and $^\spadesuit$ are the results of unsupervised SimCSE and PromptBERT respectively. 
    We report these two results from their original papers, and re-evaluate other baseline results.}
\end{table}

\section{Ablation studies} \label{app:abl}
\paragraph{Auxiliary objective in stage 2. } \label{app:aux}
    We now investigate the impact of cross entropy auxiliary objective in stage-2 training. 
    When the auxiliary loss is added, the overall objective for stage 2 has a form of 
    $\mathcal{L} = \mathcal{L}_{cl} + \alpha \cdot \mathcal{L}_{ce}$, 
    where $\alpha$ is a scalar that balances two losses. 
    Table \ref{tab:aux-loss} shows the impact of auxiliary loss in stage 2. 
    We find that for semi-supervised setting, adding auxiliary loss with $\alpha = 1\times 10^{-3}$ works the best.
    We suspect NLI data contains valuable information that benefits sentence representation learning. 
    However, for our semi-supervised method, such information gained from stage 1 can be gradually forgotten in stage 2 if no auxiliary loss is added.
    Hence, combining contrastive loss with an auxiliary cross entropy loss in stage 2 will benefit the overall training process.

\begin{table}[t]
    \centering
    \small
    \begin{tabular}{lcc}
    \toprule
      Add aux loss       & $\alpha$            & STS-B dev. \\
    \midrule
      False              & -                   & 86.2       \\
    \midrule
      True               & $1\times10^{-4}$    & 86.0       \\
                         & $1\times10^{-3}$    & \tf{86.3}  \\
                         & $1\times10^{-2}$    & 84.8       \\
                         & $1\times10^{-1}$    & 83.3       \\
    \bottomrule
    \end{tabular}
    \caption{\label{tab:aux-loss} Ablation studies of the impact of auxiliary objective during stage 2 for semi-supervised settings.}
\end{table}

\begin{figure}[t]
  \centering
  \vspace{+0.25cm}
  \includegraphics[width=1.0\linewidth]{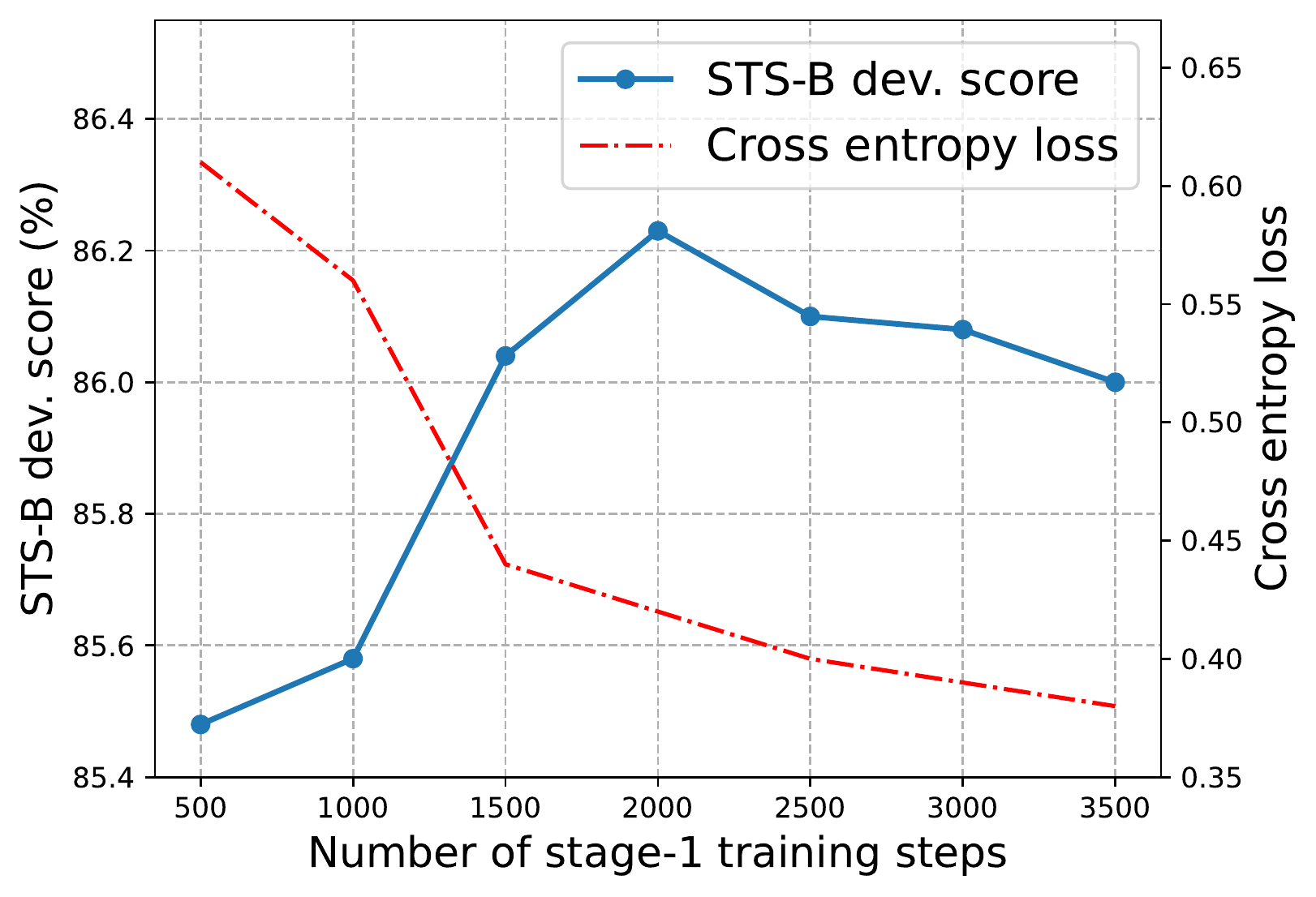}
  \caption{\label{fig:stage-1-steps} Number of stage-1 training steps V.S. STS-B development performance V.S. cross entropy loss on the training data. }
\end{figure}

\paragraph{Training steps of stage 1. }\label{app:stage1steps}
    Another hyperparameter we find crucial for the overall performance is the training steps of stage 1. 
    The importance of this hyperparameter has been elaborated in Section~\ref{sec:why-work}. 
    Here we conduct further analysis using BERT\ba \ under the semi-supervised setting by visualizing the relationship between stage-1 training steps and cross entropy loss on the training data.
    We observe that a lower stage-1 cross entropy loss does not always lead to a better overall performance. 
    From Figure~\ref{fig:stage-1-steps}, we can see that the model reaches the highest development score when the number of training steps is around 2,000, but the training loss can still be further reduced. 
    
    As we mentioned in Section~\ref{sec: diffaug},
    one of the important roles that stage-1 training performs is to make sure the prefix modules are compatible with the language model during the contrastive learning in stage 2. 
    Therefore, for our method, the compatibility between prefix and language model is more important than prefix's capability on capturing sentence relationships. 

\section{Overview of the existing parameter-efficient methods} \label{app:PE}
    In this section, we introduce two existing parameter-efficient methods that are mentioned in Section~\ref{sec:abl-pe}, i.e., Adapter~\cite{houlsby2019adapter} and LoRA~\cite{hu2021lora}.

    \paragraph{Adapter.} 
    This method proposes to insert modules with small amount of trainable parameters on each layer of the Transformer-based pre-trained language model. 
    Specifically, each module has three main components: a down-projection linear layer with parameters $W_{\text{down}} \in \mathbb{R}^{d \times r}$, a up-projection linear layer with parameters $W_{\text{up}} \in \mathbb{R}^{r \times d}$, and a non-linear activation function $f(\cdot)$, where $d$ and $r$ are the hidden states and bottleneck dimensions respectively. 
    The adpater module modifies the hidden state $\boldsymbol{h}$ in the following way:
        
        \begin{equation}
        \boldsymbol{h} \leftarrow \boldsymbol{h} + f( \boldsymbol{h} W_{\text{down}})W_{\text{up}}
        \end{equation}

    In the original implementation~\cite{houlsby2019adapter}, the adapter modules are inserted in two places on each transformer layer, i.e., one is after the self-attention layer and the other is after the feed-forward network layer.

    \paragraph{LoRA.}
    Similar to Adapter, LoRA also adds small modules, called trainable rank decomposition matrices, on each layer of the pre-trained language model. Each LoRA module is composed of two tunable weight matrices $W_{\text{down}} \in \mathbb{R}^{d \times r}$ and $W_{\text{up}} \in \mathbb{R}^{r \times d}$. Unlike Adapter, LoRA module does not contain any non-linear activation functions, and the added weight matrices are parallel to the original weight matrices. Hence, it works in the following way:
        \begin{equation}
        \boldsymbol{h} \leftarrow \boldsymbol{h} + s \cdot \boldsymbol{x} W_{\text{down}}W_{\text{up}}
        \end{equation}
    where $s \geq 1$ is a scaling factor and $\boldsymbol{x}$ represents the input. Another difference between LoRA and Adapter is that LoRA modules are only applied on the query and value projection matrices $Q$ and $V$ in the self-attention sub-layers, while the Adapter modules are attached on both the self-attention and feed-forward network sub-layers. 

\end{document}